%% file: main.tex
\documentclass{article}

\usepackage{amsmath,amssymb,amsthm}
\input{preamble}

\usepackage[preprint,nonatbib]{neurips_2023}

\usepackage[utf8]{inputenc} 
\usepackage[T1]{fontenc}    
\usepackage{hyperref}       
\usepackage{url}            
\usepackage{booktabs}       
\usepackage{amsfonts}       
\usepackage{nicefrac}       
\usepackage{microtype}      
\usepackage{xcolor}         
\usepackage{soul}
\usepackage{appendix}

\title{Duality of Bures and Shape Distances with Implications for Comparing Neural Representations}

\author{Sarah E. Harvey$^{1,2}$ \: Brett W. Larsen$^{1,2,3}$ \:  Alex H. Williams$^{1,2}$ \\
$^1$New York University, Center for Neural Science, New York, NY, 10003 \\
$^2$Flatiron Institute, Center for Computational Neuroscience, New York, NY, 10010 \\
$^3$Flatiron Institute, Center for Computational Mathematics, New York, NY, 10010 \\ 
\texttt{\{sharvey, brettlarsen, awilliams\}@flatironinstitute.org}
}

\begin{document}

\maketitle

\input{000-abstract}
\input{010-intro}
\input{020-background}
\input{021-mappings}

\input{022-similarity-matrices}

\input{030-duality}
\input{040-asymptotics}

\input{050-CKA}

\input{060-discussion}

\printbibliography
\clearpage

\begin{appendices}

\input{99-SI}
\end{appendices}

\end{document}

%% file: preamble.tex
\usepackage{amssymb}
\usepackage{amsmath}
\usepackage{amsthm}
\usepackage{amsfonts}
\usepackage{stmaryrd}
\usepackage{graphicx}

\usepackage[
backend=biber,
style=numeric,
uniquelist=false,
maxcitenames=2,
doi=false,
isbn=false,
url=false,
eprint=false,
maxbibnames=99,
sorting=nty
]{biblatex}
\AtEveryBibitem{\clearfield{month}}
\AtEveryCitekey{\clearfield{month}}
\AtEveryBibitem{\clearlist{language}}
\renewbibmacro{in:}{}
\usepackage[table,usenames,dvipsnames]{xcolor}
\usepackage[colorlinks,linktoc=all]{hyperref}
\usepackage[all]{hypcap}
\hypersetup{citecolor=Bittersweet}
\hypersetup{linkcolor=black}
\hypersetup{urlcolor=black}
\usepackage{cleveref}

\usepackage{amsthm}

\newtheorem{theorem}{Theorem}
\newtheorem{lemma}{Lemma}

\DeclareRobustCommand{\mb}[1]{\boldsymbol{#1}}

\DeclareMathOperator*{\Tr}{Tr}

\newcommand{\mbv}{\mb{v}}
\newcommand{\mbw}{\mb{w}}

\newcommand{\mbz}{\mb{z}}

\newcommand{\mbA}{\mb{A}}
\newcommand{\mbB}{\mb{B}}
\newcommand{\mbC}{\mb{C}}

\newcommand{\mbI}{\mb{I}}

\newcommand{\mbK}{\mb{K}}

\newcommand{\mbM}{\mb{M}}

\newcommand{\mbQ}{\mb{Q}}

\newcommand{\mbU}{\mb{U}}

\newcommand{\mbX}{\mb{X}}
\newcommand{\mbY}{\mb{Y}}

\newcommand{\mbLambda}{\mb{\Lambda}}

\newcommand{\mbSigma}{\mb{\Sigma}}

\newcommand{\cB}{\mathcal{B}}

\newcommand{\cF}{\mathcal{F}}
\newcommand{\cG}{\mathcal{G}}

\newcommand{\cO}{\mathcal{O}}
\newcommand{\cP}{\mathcal{P}}

\newcommand{\cS}{\mathcal{S}}

\newcommand{\cZ}{\mathcal{Z}}

\newcommand{\reals}{\mathbb{R}}
\newcommand{\expect}{\mathbb{E}}

\renewcommand{\arccos}{\cos^{-1}}

\newcommand{\pgraph}[1]{\textbf{#1.}~~}

%% file: 000-abstract.tex
\begin{abstract}
A multitude of (dis)similarity measures between neural network representations have been proposed, resulting in a fragmented research landscape.
Most of these measures fall into one of two categories.
First, measures such as linear regression, canonical correlations analysis (CCA), and shape distances, all learn explicit mappings between neural units to quantify similarity while accounting for expected invariances.
Second, measures such as representational similarity analysis (RSA), centered kernel alignment (CKA), and normalized Bures similarity (NBS) all quantify similarity in summary statistics, such as stimulus-by-stimulus kernel matrices, which are already invariant to expected symmetries.
Here, we take steps towards unifying these two broad categories of methods by observing that the cosine of the Riemannian shape distance (from category 1) is equal to NBS (from category 2).
We explore how this connection leads to new interpretations of shape distances and NBS, and draw contrasts of these measures with CKA, a popular similarity measure in the deep learning literature.

\end{abstract}

%% file: 010-intro.tex
\section{Introduction}

Quantifying similarity between neural network representations is now a well-recognized topic in computational neuroscience and deep learning~\cite{Klabunde2023,sucholutsky2023getting}.
In neuroscience, measures of representational similarity have been used to benchmark models of biological systems~\cite{Kietzmann2019,Storrs2021}, and to compare neural activity across different species~\cite{Kriegeskorte2008-man-and-monkey}.
In deep learning, they have been used to characterize learning dynamics~\cite{morcos2018insights}, model robustness~\cite{Jones2022}, and the effects of changing model architecture~\cite{Maheswaranathan2019,Nguyen2021}.

Interest in this area has sparked a proliferation of measures to quantify representational (dis)similarity including: representational similarity analysis (RSA; \cite{Kriegeskorte2008-rsa-connecting}), centered kernel alignment (CKA; \cite{Kornblith2019}), generalized shape distances \cite{Williams2021}, canonical correlations analysis~(CCA; \cite{Raghu2017}), normalized Bures similarity (NBS; \cite{Tang2020}), and the Riemannian covariance distance~\cite{shahbazi2021using}.
While all of these methods aim to quantify similar aspects of neural data, much more work is needed to formalize this intuition and to characterize the practical differences between these competing methods.

Here we develop a duality principle\footnote{Following \textcite{Atiyah2007}, we use the term \textit{duality} to broadly mean a mathematical relationship that enables ``two different points of view of looking at the same object.''} that links shape distances~\cite{kendall2009shape,Williams2021} to well-known quantities in optimal transport~\cite{Malago2018} and quantum information theory~\cite{nielsen00,quantum-fidelity-measures-review,watrous2018}.
Although similar ideas were recently described in mathematical literature on infinite-dimensional covariance operators~\cite{Masarotto2019}, these results appear thus far unnoticed within the computational neuroscience and machine learning communities.
For example, we will see that two independently proposed (dis)similarity measures---the Riemannian shape distance and NBS---are essentially equivalent to each other, 
suggesting that this duality provides way to generalize the Riemannian shape distance to cases where the networks have differing dimensionality.   
Moreover, we point out CKA and NBS have been extensively compared by quantum information theorists as different measures of similarity, CKA corresponding to the normalized Hilbert-Schmidt inner product and NBS corresponding to the \textit{fidelity} between positive semidefinite matrices with trace equal to 1~\cite{Liang2019-kh}.
An important contribution of our work is to unify these disconnected literatures with a self-contained exposition, but we also aim to demonstrate how these connections can lead to novel theoretical analysis and insights in to representational (dis)similarity.

The rest of this manuscript is organized as follows.
In \cref{sec:background}, we formalize the problem of comparing neural representations and summarize several relevant (dis)similarity measures.
In our review of past work, we classify representational (dis)similarity scores into two main categories: those that explicitly align neural dimensions, and those that quantify similarity in stimulus-by-stimulus relationships.
In \cref{sec:duality-results}, we summarize our main theoretical result linking shape distances to NBS, explicitly connecting the two categories of (dis)similarity measures.
In \cref{sec:asymptotics}, we explore the behavior of these distances when the number of stimuli ($M$) or the number of dimensions ($N$) goes to infinity.
We show that the duality between NBS and shape distance can be leveraged to understand these asymptotic regimes.
Finally, in \cref{sec:cka-comparison} we discuss how NBS and shape distances compare to CKA, a popular approach which does not enjoy the theoretical properties and interpretations discussed above.
We show numerically and analytically that the relationship between these quantities is rather loose, and we therefore do not expect them to be interchangeable in practical applications.

%% file: 020-background.tex
\section{Review of Representational (Dis)similarity Measures}
\label{sec:background}

Let $f_x : \cZ \mapsto \reals^{N_x}$ and $f_y : \cZ \mapsto \reals^{N_y}$ be two neural networks that map inputs over some domain $\cZ$ to neural activation vectors (e.g. the vector of activations produced at a hidden layer of a deep network).
Here, $N_x$ and $N_y$ respectively denote the number of neurons in each network.
This mapping from inputs to neural activations is typically considered to be deterministic (but see \cite{Duong2023} for the stochastic setting).

How similar is neural network $f_x(\cdot)$ to neural network $f_y(\cdot)$?
That is, how similar are the functions $f_x(\cdot)$ and $f_y(\cdot)$ over a representative collection of inputs $z_1, \hdots, z_M \in \cZ$?
We proceed by measuring neural responses $f_x(z_1) \dots f_x(z_M)$ and stacking them row-wise into a matrix ${\mbX \in \reals^{M \times N_x}}$.
Likewise, we form a matrix $\mbY \in \reals^{M \times N_y}$ from the second network's responses, $f_y(z_1) \dots f_y(z_M)$.
Intuitively, one can view these matrices as approximations to each network's input-output mapping over a discrete set of $M$ points.

In general, $N_x \neq N_y$,
but even if $N_x = N_y$, there is no reason we expect the raw Euclidean distance, $\Vert \mbX - \mbY \Vert_F$ to be meaningful since the neurons (columns of $\mbX$ and $\mbY$) are often labelled in arbitrary order.
Thus, we are interested in measuring a distance \textit{that is invariant to a specified set of nuisance transformations} in the representations.
For example, if we would like to ignore orthogonal transformations (including permutations of the neuron indices), we ought to develop distance functions for which $d(\mbX, \mbY) = d(\mbX, \mbY \mbQ)$ and also $d(\mbX, \mbX \mbQ) = 0$ for any orthogonal matrix $\mbQ$.
This can be formalized by defining an equivalence relation between neural responses and defining a metric over the corresponding equivalence classes~\cite{Williams2021}.

\begin{figure}\label{fig:schematic}
    \centering
    \includegraphics[width=1.0\textwidth]{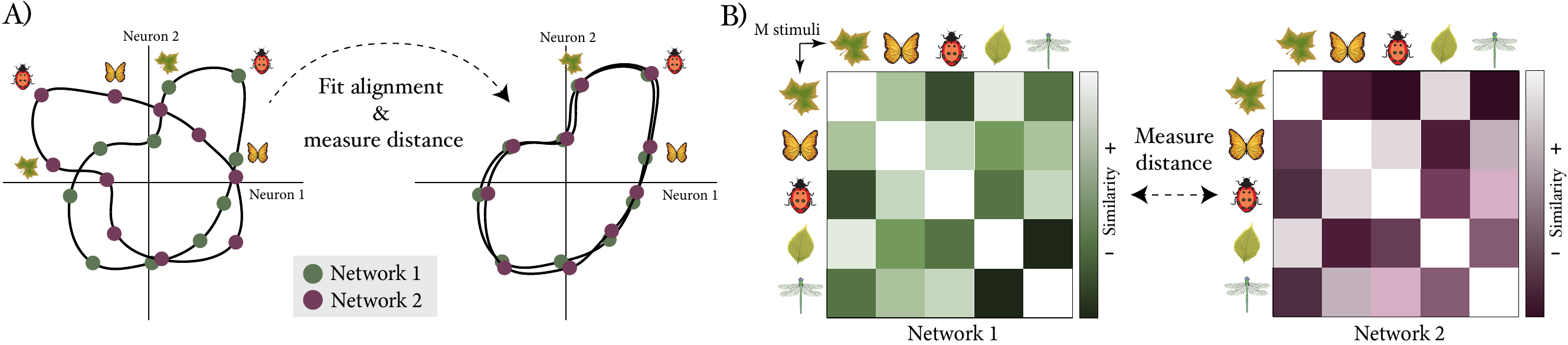}
    \caption{ We consider methods of measuring representational similarity from two broad categories: (A) alignment-based measures, which fit a mapping that aligns neural dimensions, and (B) methods that compare stimulus-by-stimulus representational similarity matrices.   }
    \label{fig:enter-label} 
\end{figure}

Existing representational (dis)similarity measures between $\mbX$ and $\mbY$ roughly fall into two broad camps: methods that learn explicit mappings to align neural dimensions, and methods that utilize stimulus-by-stimulus similarity matrices to compare across networks (\cref{fig:schematic}).
The main purpose of our paper is to provide a bridge between these approaches, and so we summarize a few primary examples below.
A comprehensive review of these methods is far beyond the scope of this paper, but we point the reader to \textcite{Klabunde2023,sucholutsky2023getting} as useful papers to cross-reference.
Indeed, \textcite{Klabunde2023} enumerate over 30 methods to quantify representational (dis)similarity, showing the strong need for organizing theoretical principles to relate and unify these approaches.

%% file: 021-mappings.tex
\subsection{(Dis)similarity measures that transform or align neural dimensions}
\label{subsec:mappings}

Here we review the first major category of representational (dis)similarity measures.
Recall that the main challenge is that raw distances, such as ${\Vert \mbX - \mbY \Vert_F}$, are typically meaningless due to nuisance transformations.
To overcome this, one option is to optimize alignment functions $g_x : \reals^{N_x} \mapsto \cS$ and $g_y : \reals^{N_y} \mapsto \cS$ which respectively map the rows of $\mbX$ and $\mbY$ into a common space $\cS$.
Then, given a dissimilarity measure ${d : \cS \times \cS \mapsto \reals}$ we can optimize the alignment:
\begin{equation}
\label{eq:mapping-dissimilarity-measures}
\underset{g_x, g_y}{\text{minimize}} ~~ d(g_x(\mbX), g_y(\mbY)) \quad \text{subject to}~~g_x \in \cG_x ~, ~ g_y \in \cG_y
\end{equation}
where $\cG_x$ and $\cG_y$ represent the class of permitted alignment functions for $\mbX$ and $\mbY$, respectively.
Note that we are using $g_x(\mbX)$ and $g_y(\mbY)$ to denote the row-wise application of $g_x(\cdot)$ and $g_y(\cdot)$ to matrix-valued inputs.

Intuitively, the minimal value attained in \cref{eq:mapping-dissimilarity-measures} will be invariant to nuisance transformations that are contained within $\cG_x$ or $\cG_y$.
For example, \textbf{linear regression} is a popular method to quantify similarity between artificial and biological neural networks~\cite{schrimpf2018brain}, and can be viewed as a special case of \cref{eq:mapping-dissimilarity-measures}.
Specifically, to predict $\mbY$ from $\mbX$, we would choose $\cS = \reals^{N_y}$, constrain $g_y(\cdot)$ to be the identity mapping, and minimize $g_x(\cdot)$ over the set of affine mappings from ${\reals^{N_x} \mapsto \reals^{N_y}}$.
Importantly, linear regression does not produce a symmetric notion of (dis)similarity: predicting $\mbY$ from $\mbX$ will produce a different result than predicting $\mbX$ from $\mbY$.
Variants such as \textbf{canonical correlations analysis (CCA)} provide symmetric, affine-invariant measures of representational similarity, and have been used in deep learning and in neuroscience~\cite{sussillo2015neural,Raghu2017,gallego2020long}.
To see that CCA is a special case of \cref{eq:mapping-dissimilarity-measures}, we choose $\cS = \reals^{n}$ where $n = \min(N_x, N_y)$ and minimize ${-1 \cdot \Tr[g_x(\mbX)^\top g_y(\mbY)]}$ subject to ${g_x(\mbX)^\top g_x(\mbX) = g_y(\mbY)^\top g_y(\mbY) = \mbI}$.
Intuitively, CCA finds mappings of $\mbX$ and $\mbY$ into a common space $\reals^n$ which maximize correlations along orthogonal dimensions.
Other methods like \textbf{permutation matching} of individual neurons \cite{Li2015}, can also be viewed as special cases of \cref{eq:mapping-dissimilarity-measures}.

In summary, the examples above show that \cref{eq:mapping-dissimilarity-measures} captures a broad variety of existing approaches.
We now review another example, \textbf{shape distances}~\cite{kendall2009shape}, which feature prominently into our main narrative.
\textcite{Kendall1977-ty} defined shape as the structure left by a set of $M$ \textit{landmark} points in $\reals^N$ after rotations, translations, and isotropic scalings are ignored. 
These $M$ landmark points in our context become the collection of $M$ inputs to each network over which the representational similarity will be measured. 
Unlike traditional shape theory, we will additionally consider reflections as in $\reals^N$ as nuisance transformations, since a permutation of neuron labels (which we argued above is typically arbitrary) can require a reflection.

Assuming that $\mbX$ and $\mbY$ are $M \times N$ matrices, we can define their angular distance:
\begin{equation}
\label{eq:angular-distance}
\theta(\mbX, \mbY) = \arccos \left ( \frac{\Tr [\mbX^\top \mbY ]}{\sqrt{\Tr [ \mbX^\top \mbX  ] \Tr [ \mbY^\top \mbY  ] } } \right ).
\end{equation}
which generalizes the elementary formula for the angle between two vectors.
One can then define the \textbf{Riemannian shape distance} as the length of the shortest geodesic path between two shapes, and show that this is given by~\cite{kendall2009shape}:
\begin{equation}
\label{eq:riemannian-shape-distance}
\theta^*(\mbX, \mbY) = \min_{\mbQ^\top \mbQ = \mbI} \theta(\mbC \mbX, \mbC \mbY \mbQ),
\end{equation}
where $\mbC = (\mbI - \tfrac{1}{M} \mb{1}\mb{1}^\top)$
is the \textit{centering matrix} ($\mb{1}\mb{1}^\top$ is an $M\times M$ matrix of all ones).
One may check that the columns of $\mbC \mbX$ and $\mbC \mbY$ are mean-centered, and that $\mbC$ is a symmetric, idempotent matrix with $\mbC^\top \mbC = \mbC^2 = \mbC$.

Closely related to the Riemannian shape distance is a quantity called the \textbf{Procrustes size-and-shape distance} ~\cite{kendall2009shape} (which, for brevity, we will simply call the Procrustes distance):
\begin{equation}
\label{eq:procrustes-shape-distance}
\mathcal{P}(\mbX, \mbY) = \min_{\mbQ^\top \mbQ = \mbI} \Vert \mbC \mbX - \mbC \mbY \mbQ \Vert_F
\end{equation}
After expanding \cref{eq:procrustes-shape-distance} and rearranging, we can see that the cosine of the Riemannian shape distance is related to the Procrustes distance between two matrices $\mbX$ and $\mbY$ normalized by their centered Frobenius norms:
\begin{equation}\label{eq:riemann-procrustes}
    \cos{\theta^{*}(\mbX,\mbY) } = 1 - \frac{1}{2}\mathcal{P}^2\bigg{(} \frac{\mbX}{\| \mbC \mbX \|_{F}}, \frac{\mbY}{\| \mbC \mbY \|_{F}}  \bigg{)}.
\end{equation}
It is straightforward to check that both Riemannian shape distance and Procrustes distance are invariant to translations and orthogonal transformations.
The Riemannian shape distance is additionally invariant to isotropic scalings.
Again, both of these distances are special cases of \cref{eq:mapping-dissimilarity-measures}.

We note that both shape distances are symmetric and satisfy the triangle inequality.
That is, for any triplet of configuration matrices $\mbX, \mbY, \mbM \in \reals^{M \times N}$ we have $\theta^*(\mbX, \mbY) = \theta^*(\mbY, \mbX)$ and $\theta^*(\mbX, \mbY) \leq \theta^*(\mbX, \mbM) + \theta^*(\mbM, \mbY)$, and likewise for $\cP$.
Formally, this means that $\theta^*$ and $\cP$ define metric spaces over the equivalence classes defined by their nuisance transformations.
\textcite{Williams2021} argued that these properties were advantageous for downstream analyses such as nearest-neighbor regression and clustering methods that leverage metric space structure.

\pgraph{Comparing shapes of unequal dimension}
Importantly, the definitions of shape distance in \cref{eq:riemannian-shape-distance,eq:procrustes-shape-distance} assume that we are comparing networks of equal size---i.e., that $N_x = N_y = N$.
This corresponds to the setting of traditional shape theory, but is not an ideal assumption for our application since we often desire to compare representations across networks with different sizes or different numbers of experimentally measured neurons.
\textcite{Williams2021} proposed procedures to  either use PCA or zero padding to embed all networks into a common dimension.
We will show these procedures are unnecessary since we can reformulate the shape distances in terms of:
\begin{equation}
\label{eq:covariance-def}
\mbSigma_X = \mbX^\top \mbC \mbX \quad \text{,} \quad \mbSigma_Y = \mbY^\top \mbC \mbY \quad \text{and} \quad \mbSigma_{XY} = \mbX^\top \mbC \mbY
\end{equation}
which are the covariance and cross-covariance matrices describing similarity between pairs of neural tuning functions within and across networks.
Specifically, let us define:
\begin{align}
\label{eq:riemannian-alt}
\theta^*(\mbX, \mbY) &= \arccos \left [ \frac{ \Vert \mbSigma_{XY} \Vert_*}{\sqrt{\Tr[\mbSigma_{X}] \Tr[\mbSigma_{Y}]}} \right ] \\[.3em]
\label{eq:procrustes-alt}
\mathcal{P}(\mbX, \mbY) &= \sqrt{\Tr[\mbSigma_X] + \Tr[\mbSigma_Y] - 2 \Vert \mbSigma_{XY} \Vert_*}
\end{align}
where $\Vert \cdot \Vert_*$ denotes the nuclear matrix norm (also called the Schatten 1-norm), which is given by the sum of a matrix's singular values.
Our main claim is the following:
\begin{lemma}
\label{lemma:procrustes-covariance}
When $N_x = N_y = N$, the definitions of $\theta^*$ and $\cP$ given in \cref{eq:riemannian-shape-distance,eq:procrustes-shape-distance} are equivalent to the definitions of $\theta^*$ and $\cP$ given in \cref{eq:riemannian-alt,eq:procrustes-alt}.
\end{lemma}
which follows immediately from a well-known result of \textcite{Schonemann1966} (Appendix~\ref{subsec:proof-of-lemma-1}).

\Cref{lemma:procrustes-covariance} essentially shows that \cref{eq:riemannian-alt,eq:procrustes-alt} are reasonable generalizations of shape distance that are well-defined when $N_x \neq N_y$.
Although it may not be immediately obvious, we will see (due to \cref{theorem:duality} in \cref{sec:duality-results}) that the new definitions of shape distance in \cref{eq:riemannian-alt,eq:procrustes-alt} continue to satisfy the triangle inequality, even when comparing networks with different sizes.
The geometric intuition underlying shape distances---that of fitting a rotational alignment between two neural activation spaces---also carries over.
Specifically, let us assume $N_x \leq N_y$ (without loss of generality).
Then, we can isometrically embed the lower-dimensional representations into $\reals^{N_y}$ by, for example, appending $N_y - N_x$ columns of zeros to $\mbX$.
Then, we compute the shape distance as before, which involves finding the optimal orthogonal transformation in $N_y$ dimensions to match the landmark points.

%% file: 022-similarity-matrices.tex
\subsection{(Dis)similarity measures that quantify stimulus-by-stimulus relationships}
\label{subsec:similarity-matrices}

Recall that the (dis)similarity measures summarized above relied on learning explicit alignment transformations on the neural activation space, $\reals^{N}$ for a population of $N$ neurons.
We now turn to review the second, alternative category of measures, which avoid the need to fit any alignment.
Instead, these methods quantify representational similarity by comparing summary statistics that are already invariant to nuisance transformations.

For instance, given $N$-dimensional network responses to $M$ sampled inputs, we can compute a \textbf{representational dissimilarity matrix (RDM)} ~\cite{Kriegeskorte2008-rsa-connecting}: an $M \times M$ matrix of Euclidean distances between all pairs of evoked responses within the $N$-dimensional response space.
Intuitively, RDMs encode a rich geometric summary of the network's representation that is invariant to rotations and translations of the neural activation space.
In fact, since the Procrustes distance is invariant to translations, it is easy to show that $\mbX$ and $\mbY$ have the same RDM if and only if the size-and-shape Procrustes distance between $\mbX$ and $\mbY$ is zero.
This already hints at deeper connections, which we reveal in \cref{sec:duality-results}.

Quantifying similarity of RDMs across networks is common in cognitive/systems neuroscience, where the approach is broadly referred to as \textbf{representational similarity analysis (RSA)}~\cite{Kriegeskorte2008-rsa-connecting}.
Many variants of RSA use different approaches to compute the within-network RDMs (e.g. Mahalanobis vs. Euclidean distance) or different measures to compare two RDMs (e.g. Pearson or Spearman correlation scores)~\cite{nili2014toolbox,walther2016reliability,Schutt2023}.
But all of these variants conceptually share the same core approach.

An alternative to computing RDMs is to compute $M \times M$ \textit{kernel matrices}, which use a positive definite kernel function~\cite{scholkopf2002learning} to compute a similarity score between all pairs of evoked network responses.
Specifically, we will focus on \textit{centered linear kernel matrices} (with centering matrix $\mbC$ defined as in the previous section), as they are the most popular in practice:
\begin{equation}
\label{eq:centered-kernel-matrices}
\mbK_X = \mbC \mbX \mbX^T \mbC
\quad \text{and} \quad 
\mbK_Y = \mbC \mbY \mbY^T \mbC ~.
\end{equation}
Note that $\mbK_X$ and $\mbK_Y$ are simply covariance matrices across stimuli; they are natural counterparts to the covariance matrices across neurons, $\mbSigma_X$ and $\mbSigma_Y$ defined in \cref{eq:covariance-def}.
Intuitively, $\mbK_X$ and $\mbK_Y$ hold all the information necessary to compute an RDM since the squared Euclidean distance between stimulus $i$ and $j$ is given by:
${[\mbK_X]_{ii} + [\mbK_X]_{jj} - [\mbK_X]_{ij}^2}$.
Like RDMs, the kernel matrices are invariant to rotations, reflections, and translations.
An influential paper by \textcite{Kornblith2019} proposed \textbf{centered kernel alignment (CKA)} \cite{Cristianini2001,Cortes2012-bp} as a measure of similarity between kernel matrices:
\begin{equation}
\label{eq:cka-definition}
\text{CKA}(\mbK_X, \mbK_Y) = \frac{\Tr[\mbK_X \mbK_Y]}{\sqrt{\Tr[\mbK_X^2] \Tr[\mbK_Y^2]}}
\end{equation}
which is the cosine of the angle between $\mbK_X$ and $\mbK_Y$ (see eq.~\ref{eq:angular-distance}).
\textcite{shahbazi2021using} pointed out that CKA does not exploit the fact that $\mbK_X$ and $\mbK_Y$ are positive semidefinite (PSD) matrices, and they propose an alternative metric based on the \textbf{Riemannian distance on PSD matrices}.
Yet another measure on the linear kernel matrices is the \textbf{normalized Bures similarity (NBS)}, defined as \cite{Muzellec2018,Tang2020}:
\begin{equation}
\label{eq:nbs-definition}
\text{NBS}(\mbK_X, \mbK_Y) = \frac{\cF(\mbK_X,\mbK_Y)}{\sqrt{ \Tr[\mbK_X] \Tr[\mbK_Y] }} ~
\end{equation}
with 
\begin{equation}\label{eq:fid-definition}
\cF(\mbK_X,\mbK_Y) = \Tr[ (\mbK_X^{1/2} \mbK_Y \mbK_X^{1/2})^{1/2}].  
\end{equation}
The quantity $\cF(\mbK_X,\mbK_Y)$ is known as the \emph{fidelity} and is used extensively in quantum information theory as a measure of the distinguishability of quantum states~\cite{quantum-fidelity-measures-review,watrous2018}.
We will also make use of a related quantity known as the \textbf{Bures distance} on PSD matrices~\cite{Bhatia2019}:
\begin{equation}
\label{eq:bures-distance-definition}
\cB(\mbK_X, \mbK_Y) = \sqrt{\Tr[\mbK_X] + \Tr[\mbK_Y] - 2 \Tr \left [ \left (\mbK_X^{1/2} \mbK_Y \mbK_X^{1/2} \right )^{1/2} \right] } .
\end{equation}
It is well-known that the Bures distance is equal to the 2-Wasserstein distance between two mean-centered normal distributions with covariances given by $\mbK_X$ and $\mbK_Y$~(\cite{Peyre2019}, Remark 2.30).
Thus, one can interpret $\cB(\mbK_X, \mbK_Y)$ as the cost of optimally transporting mass between two normal densities.
This connection could allow one to exploit the large collection of existing knowledge of optimal transportation, as in \cite{Masarotto2019}, although this is beyond the scope of the present work.

%% file: 030-duality.tex
\section{Duality of Shape and Bures Distances}
\label{sec:duality-results}

In \cref{sec:background}, we saw that many measures of representational (dis)similarity either identify an explicit mapping between neural dimensions or directly compare stimulus-by-stimulus (dis)similarities.
Each perspective has its own conceptual appeal.
The former encourages us to reason about geometric features in the space of neural activations, such as curvature or tangling of manifold structure which feature prominently in theories of neural computation \cite{russo2018motor,henaff2021primary,harrington2023exploring}.
The latter avoids the need to align neural axes, and connects to a rich literature in psychology that leverages pairwise similarity judgements to interrogate the structure of cognition \cite{beals1968foundations,edelman1998representation}.
Our sense is that many researchers at the intersection of neuroscience, cognitive science, and interpretable deep learning tend to develop a personal preference for one perspective over the other.
But are these approaches really so distinct?

We now turn to one of our main results, which highlights a specific case where these two perspectives produce the same quantitative result---an example of a \textit{duality}~\cite{Atiyah2007}.
Specifically, the theorem below states that the Procrustes distance ($\cP$, eqs.~\ref{eq:procrustes-shape-distance}~and~\ref{eq:procrustes-alt}) is equal to the Bures distance between linear kernel matrices ($\cB$, eq.~\ref{eq:bures-distance-definition}).
Furthermore, the normalized Bures similarity (NBS, eq.~\ref{eq:nbs-definition}) is equal to the cosine of the Riemannian shape distance ($\theta^*$, eqs.~\ref{eq:riemannian-shape-distance}~and~\ref{eq:riemannian-alt}).

\begin{theorem}
\label{theorem:duality}
Let $\mbK_X$ and $\mbK_Y$ be centered linear kernel matrices as in \cref{eq:centered-kernel-matrices}. Then,
\begin{equation}
\label{eq:theorem-equivalence-of-bures-procrustes}
\cB(\mbK_X, \mbK_Y) = \mathcal{P}(\mbX, \mbY)
\end{equation}
and furthermore,
\begin{equation}
\label{eq:theorem-equivalence-of-nbs-shape-dist}
\text{NBS}(\mbK_X, \mbK_Y) = \cos \theta^*(\mbX, \mbY) ~.
\end{equation}
\end{theorem}

The proof follows from the fact that for centered linear kernel matrices, $\Tr[\mbK_X] = \Tr[\mbSigma_X]$ and $\Tr[\mbK_Y] = \Tr[\mbSigma_Y]$, and the following lemma:

\begin{lemma}
\label{lemma:fidelity-nuclear-norm}
For centered linear kernel matrices, $\cF(\mbK_X, \mbK_Y) = \Vert \mbSigma_{XY} \Vert_*$.
\end{lemma}

It is easy to see that the lemma implies the theorem. E.g., from the definitions in \cref{eq:procrustes-alt,eq:bures-distance-definition}:
\begin{align*}
\cB^2(\mbK_X, \mbK_Y) &= \Tr[\mbK_X] + \Tr[\mbK_Y] - 2 \cF(\mbK_X, \mbK_Y) \\
\mathcal{P}^2(\mbX, \mbY) &= \Tr[\mbSigma_X] + \Tr[\mbSigma_Y] - 2 \Vert \mbSigma_{XY} \Vert_*
\end{align*}
So \cref{eq:theorem-equivalence-of-bures-procrustes} follows from observing that the three terms on the right hand sides above are equal due to \cref{lemma:fidelity-nuclear-norm}.
A similar argument shows that \cref{eq:theorem-equivalence-of-nbs-shape-dist} follows from lemma 2 as well.
Thus, all that remains is to prove lemma 2, which we do in Appendix~\ref{subsec:proof-fidelity-nuclear-norm}.

The proof of \cref{theorem:duality} is straightforward, and somewhat similar results have already appeared in mathematical literature~\cite{Masarotto2019}.
However, this result may have gone unnoticed by researchers at the intersection of machine learning and neuroscience because prior similar statements have appeared in a technical literature focused on distinct problems.

\Cref{theorem:duality} enables us to draw upon an extensive literature to theoretically characterize shape/Bures distances.
For example, it is well known that $\cB$ and $\arccos \text{NBS}$ both satisfy the criteria of a metric space, including the triangle inequality~\cite{nielsen00}.
Thus, we can immediately conclude that the generalized definitions of Procrustes and Riemannian shape distance in \cref{eq:riemannian-alt,eq:procrustes-alt} are also metrics, even though most classical work on shape theory does not typically consider datasets with unequal dimensions ($N_x \neq N_y$).

%% file: 040-asymptotics.tex
\vspace{-2mm}
\section{Asymptotic Analysis of Shape/Bures Distances}
\label{sec:asymptotics}
\vspace{-1mm}

In the previous section, we provide a concrete link between two previously disconnected perspectives of representaitonal (dis)similarity.
Beyond conceptual appeal, does this advance unlock any practical benefits for future research?
To demonstrate the utility of our result, we investigate how shape/Bures distance changes as more inputs are sampled ($M \rightarrow \infty$) and as the size of the neural population increases ($N \rightarrow \infty$).
As shown below, both of these regimes are of interest to researchers in neuroscience and deep learning, and the duality established in \cref{theorem:duality} enables immediate insights.

Before proceeding, we must introduce normalization factors into our definitions of Procrustes and Bures distances, since $\cP(\mbX, \mbY)$ and $\cB(\mbK_X, \mbK_Y)$ will diverge as ${N,M \rightarrow \infty}$.
Thus, we define the normalized Procrustes distance, $\rho$, and normalized Bures distance, $b$, as:
\begin{equation}
\rho(\mbX, \mbY) = \tfrac{1}{\sqrt{NM}} \cP(\mbX, \mbY) \quad \text{and} \quad  b(\mbK_X, \mbK_Y) = \tfrac{1}{\sqrt{NM}} \cB(\mbK_X, \mbK_Y)
\end{equation}
Note that we have assumed that $N_x = N_y = N$ (i.e. the networks have the same number of neurons) since we are interested in taking $N \rightarrow \infty$ anyways.
The motivation behind the normalizing factor of $1 / \sqrt{M N}$ will be made clear by the discussion below.

\pgraph{Interpretation as $M \rightarrow \infty$} Thus far, we have considered network representations of $M$ input stimuli. In most cases, these inputs are viewed as from a broader input distribution. For example, to compare visual representations, we input $M$ random natural images into a pair of networks and measure their similarity. As $M \rightarrow \infty$ we would desire that $\rho(\mbX, \mbY)$ and $b(\mbK_X, \mbK_Y)$ converge to a constant value that reflects the underlying structure of the input distribution.
But in this limit $\mbK_X$ and $\mbK_Y$ become, loosely speaking, infinite-dimensional matrices.
Thus it may not be immediately obvious how to interpret $b(\mbK_X, \mbK_Y)$ without tools from functional analysis.  

On the other hand, due to the law of large numbers, we have in this limit that:
\begin{equation}
\label{eq:lln-neuron-covariance}
\tfrac{1}{M} \mbSigma_X \rightarrow \mbSigma_X^*
\hspace{2em}
\tfrac{1}{M} \mbSigma_Y \rightarrow \mbSigma_Y^*
\hspace{2em}
\tfrac{1}{M} \mbSigma_{XY} \rightarrow \mbSigma_{XY}^*
\end{equation}
where $\mbSigma_X^*$, $\mbSigma_Y^*$, and $\mbSigma_{XY}^*$ are the true covariances and cross-covariances across neurons.
Thus,
\begin{equation}
\label{eq:large-m-limit}
\lim_{M \rightarrow \infty} \rho(\mbX, \mbY) = \sqrt{\tfrac{1}{N}} \cdot \sqrt{\Tr[\mbSigma_X^*] + \Tr[\mbSigma_Y^*] - 2 \Vert \mbSigma_{XY}^* \Vert_* }
\end{equation}
which is in some sense the ``true'' Procrustes distance we are aiming to approximate, pre-multiplied by a factor of $1/\sqrt{N}$.
\Cref{theorem:duality} allows us to immediately conclude that $b(\mbK_X, \mbK_Y)$ converges to the same value as $\rho(\mbX, \mbY)$ in this limit.

\pgraph{Interpretation as $N \rightarrow \infty$} Experimental neuroscientists often record a small random sample of $N$ neurons (e.g. 100-1000 cells) from brain regions that are much larger, often by several orders of magnitude.
How large do we need $N$ to be in order to achieve a good estimate of the ``true'' representational (dis)similarity~\cite{kriegeskorte2016-jl-lemma-review}?
In analogy to our logic above, the ``true'' (dis)similarity is given by considering the limit that $N \rightarrow \infty$.
This limiting regime is also of interest to the theory of deep learning, in which one often deals with networks with ``infinitely wide'' layers~\cite{Matthews2018,jacot2018neural,lee2018deep}.

In the limit that $N \rightarrow \infty$, shape distances become hard to conceptualize without leveraging advanced mathematics.
Conceptually, one can imagine fitting an infinite-dimensional rotation matrix that aligns the neural axes, or else calculating \cref{eq:procrustes-alt} on infinite-dimensional covariances.
On the other hand, in analogy to \cref{eq:lln-neuron-covariance}, we have:
\begin{equation}
\tfrac{1}{N} \mbK_X \rightarrow \mbK_X^*
\hspace{4em}
\tfrac{1}{N} \mbK_Y \rightarrow \mbK_Y^*
\end{equation}
in the limit that $N \rightarrow \infty$ due to the law of large numbers.
Intuitively, $\mbK^*_X$ and $\mbK^*_Y$ are the ``true'' covariances describing the expected similarity between pairs of stimuli across the fully observed neural population.
Further, it is easy to verify that the normalized Bures distance converges to the appropriate value, up to a factor of $1 / \sqrt{M}$. That is,
\begin{equation}
\label{eq:large-n-limit}
\lim_{N \rightarrow \infty} b(\mbX, \mbY) = \sqrt{\tfrac{1}{M}} \cdot \sqrt{\Tr[\mbK_X^*] + \Tr[\mbK_Y^*] - 2 \cF(\mbK_X^*, \mbK_Y^*) }
\end{equation}
\Cref{theorem:duality} allows us to immediately conclude that $\rho(\mbX, \mbY)$ also converges to this value in this limit.

\pgraph{Interpretation as $M \rightarrow \infty$ and $N \rightarrow \infty$} In summary, \cref{theorem:duality} makes it easy to see that normalized shape and Bures distances converge to reasonable values when either $M \rightarrow \infty$ or $N \rightarrow \infty$.
Deriving this insight only required us to leverage a basic law of large numbers---namely, that empirical covariance matrices converge to the true covariance in the limit of infinite samples.

An obvious question is whether the shape/Bures distances converge to reasonable quantities when \textit{both} $M$ and $N$ are taken to infinity.
A rigorous analysis of this scenario requires a more delicate approach that leverages concepts from functional analysis.
Nonetheless, it can be shown that as $M, N \rightarrow \infty$, the Bures distance between covariance matrices converges to the Bures distance between an associated positive semidefinite 
covariance operator \cite{pigoli2014,mallasto2017}.

%% file: 050-CKA.tex
\vspace{-1mm}
\section{Theoretical and Numerical Comparisons with CKA}
\label{sec:cka-comparison}
\vspace{-1mm}

We have seen that shape and Bures distances enjoy a special duality.
Namely, they can be interpreted as the distance found after optimally aligning the neural activation spaces (see eqs.~\ref{eq:riemannian-shape-distance} and \ref{eq:procrustes-shape-distance}) or a direct distance between stimulus-by-stimulus kernel matrices.
This special property does not appear to be shared by other representational (dis)similarity measures, which are typically compatible with only one of these perspectives.
However, the shape and Bures distances only represent a very small fraction of a much larger landscape of (dis)similarity measures, which we surveyed briefly in \cref{sec:background}.
Do the shape and Bures distances meaningfully differ from these alternative approaches?

In this section, we investigate the relationship between NBS and CKA~\cite{Cortes2012-bp,Cristianini2001,Kornblith2019}, a particularly popular approach in the deep learning literature.
By comparing their definitions in \cref{eq:cka-definition,eq:nbs-definition}, one may guess that CKA is closely related to NBS (and therefore also to Riemannian shape distance by \cref{theorem:duality}).
However, we will show that CKA scores between networks can differ substantially (e.g. two- to three-fold) from NBS scores.
We also derive upper and lower bounds that relate CKA and NBS; an exercise which confirms their rather loose relationship.
Overall, we conclude that one should not expect CKA and NBS to behave similarly in practical scenarios.

\subsection{Relationship between CKA and Euclidean geometry}

We begin our comparison by noting a relationship between CKA and Euclidean distance:
\begin{equation}
\label{eq:cka-euclidean-dist-connection}
\text{CKA}(\mbK_X,\mbK_Y) = 1 - \frac{1}{2} \bigg{\|} \frac{\mbK_X}{\| \mbK_X \|_F}  - \frac{\mbK_Y}{\| \mbK_Y \|_F}  \bigg{\|}_{F}^{2}.
\end{equation}
Likewise, by noting that $\text{NBS}(\mbK_X, \mbK_Y) = \cF(\mbK_X/\Tr{\mbK_X} ,\mbK_Y/\Tr{\mbK_Y})$ and rearranging the definition of Bures distance, we observe that NBS has an analogous relationship to Bures distance:
\begin{equation}
\label{eq:nbs-bures-dist-connection}
\text{NBS}(\mbK_X, \mbK_Y) = 1 - \frac{1}{2} \mathcal{B}^2\Big{(} \frac{\mbK_X}{\Tr \mbK_X},  \frac{\mbK_Y}{\Tr \mbK_Y} \Big{)}
\end{equation}
which can be compared with \cref{eq:riemann-procrustes}.
Thus, the central conceptual difference between CKA and NBS is the choice of metric on the space of PSD matrices.
Intuitively, measuring similarity with CKA (instead of NBS) is akin to measuring distance using a Euclidean (instead of Bures) geometry on PSD matrices.

Many previous works have argued that using a Euclidean geometry to compare or estimate PSD matrices is suboptimal for certain analyses~\cite{Arsigny2007,Dryden2009}, including recent work by \textcite{shahbazi2021using} in the context of comparing neural representations.
To gain some intuition, consider the problem of interpolating between two kernel matrices. 
In a Euclidean geometry, one obtains $\alpha\mbK_X + (1 - \alpha) \mbK_Y$ which is PSD if $0 \leq \alpha \leq 1$.
However, if one extrapolates beyond these bounds (e.g. by choosing $\alpha > 1$), the resulting matrix may contain negative eigenvalues.
Such extrapolations are used when CKA is optimized by gradient descent, as done in several prior works~\cite{davari2023reliability,dapello2023aligning}.

The problems mentioned above do not arise if one extrapolates along geodesics defined by the Bures distance.
Indeed, the Bures distance between a PSD matrix and a matrix with negative eigenvalues is not well-defined.
However, it is not the main point of this paper to contend that Bures geometry is inherently superior.
One can show that the topologies induced by the Euclidean and Bures distances coincide (see Lemma 3.2 in \cite{Thanwerdas2023}).
Determining whether the Euclidean geometry is ``good enough'' for a particular application should be considered carefully on a case-by-case basis.

\begin{figure}\label{fig:NBS-CKA}
    \centering
    \includegraphics[width=1.0\textwidth]{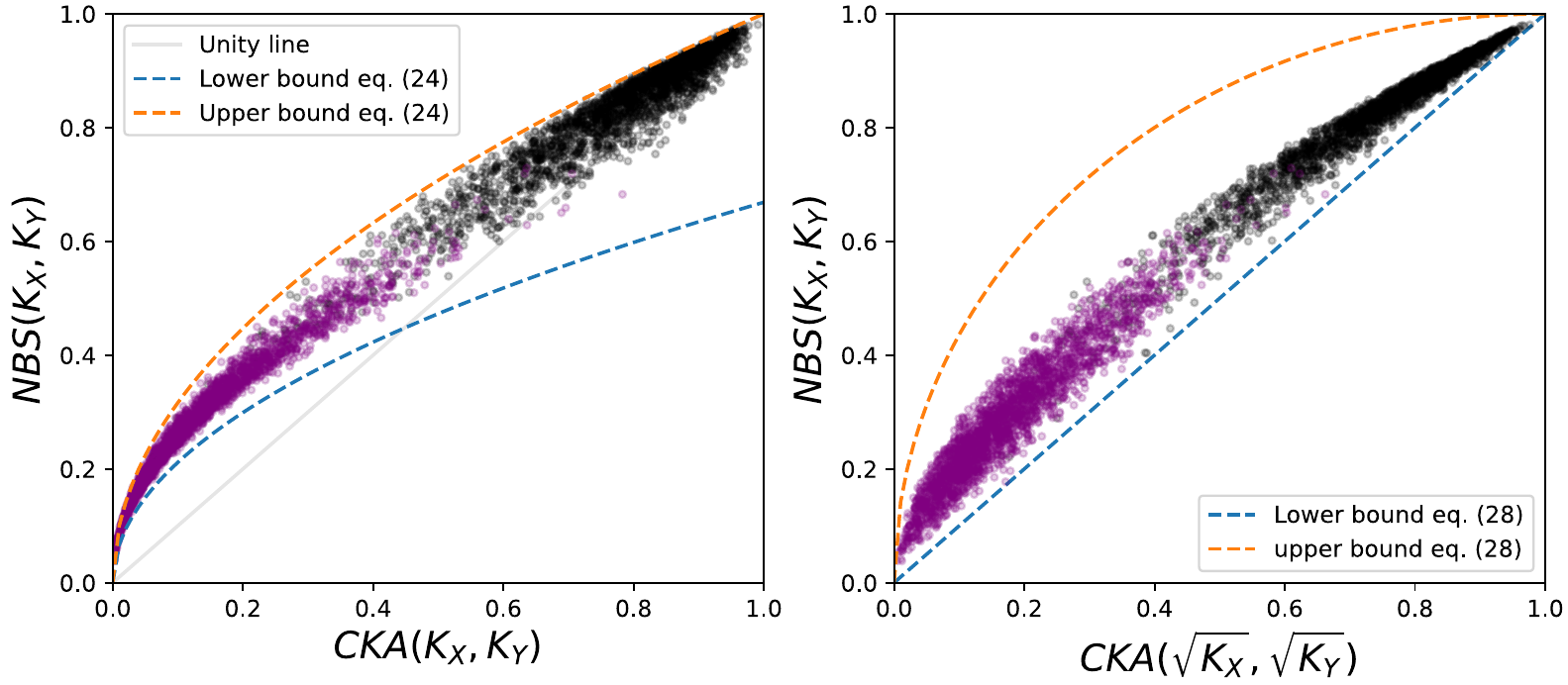}
    \caption{Comparing CKA and NBS. Purple points represent the similarity between pairs of matrices generated by sampling two Wishart distributions, $\sqrt{\mbK_X} \sim W_{10}(\mbI,1)$ and $\sqrt{\mbK_Y} \sim W_{10}(\mbI,5)$.  Black points are generated by sampling $\sqrt{\mbK_X} \sim W_{10}(\mbI,1)$ and setting $\sqrt{\mbK_Y} = \sqrt{\mbK_X} + \epsilon$, where $\epsilon \sim W_{10}(\mbI,4)$. (a) CKA bounds NBS within an envelope determined by the matrix ranks; see \cref{eq:NBS-envel}. (b) CKA($\sqrt{\mbK_X}, \sqrt{\mbK_Y}$) bounds NBS($\mbK_X,\mbK_Y$) with inequalities given by \cref{eq:sqCKAbound}.         }
    \label{fig:enter-label} 
\end{figure}

\subsection{Upper and lower bounds on NBS in terms of CKA}
\label{subsec:cka-nbs-bounds-1}

Having discussed a central conceptual difference between NBS and CKA, we turn our attention to a more practical question: How big are the potential discrepancies between CKA and NBS?
\Cref{fig:NBS-CKA} shows NBS plotted against CKA for randomly sampled pairs of PSD matrices.  
This figure suggests that, while there is not a one-to-one relationship between the two quantities, the two similarity measures constrain each other to an allowed envelope.  
We can derive bounds on this envelope by expressing squared NBS and CKA in terms of matrix norms:
\begin{equation}
    \label{eq:CKA-NBS-matrix-norms}
    \text{NBS}(\mbK_X,\mbK_Y)^2 = \frac{\| \mbSigma_{XY} \|_{*}^{2}}{\| \mbK_X \|_{*}  \| \mbK_Y \|_{*}   }, \:\: \text{CKA}(\mbK_X,\mbK_Y) = \frac{\| \mbSigma_{XY} \|_{F}^{2}}{\| \mbK_X \|_{F}  \| \mbK_Y \|_{F} } 
\end{equation}
(Note that we have exploited \cref{theorem:duality} to reformulate the numerator of NBS.)
The elementary matrix norm inequalities $\| \mbA \|_{F} \leq  \| \mbA \|_{*} \leq \sqrt{\text{rank}(\mbA)} \| \mbA \|_{F}$
and the subsequent observation $\Tr[\mbK_X \mbK_Y] \leq \mathcal{F}(\mbK_X, \mbK_Y)^2$ lead us to the following envelope (setting $r(\mbA) = \text{rank}(\mbA))$:
\begin{equation}\label{eq:NBS-envel}
    \frac{\text{CKA}(\mbK_X,\mbK_Y)}{\sqrt{r(\mbK_X) r(\mbK_Y)}}  \leq \text{NBS}(\mbK_X,\mbK_Y)^2  \leq \min[r(\mbK_X),r(\mbK_Y)]\, \text{CKA}(\mbK_X,\mbK_Y)
\end{equation}
Both of these bounds are saturated when $\text{rank}(\mbK_X) = \text{rank}(\mbK_Y) = 1$.
\Cref{fig:NBS-CKA} (a) demonstrates that while NBS is bound to an envelope by CKA set by the matrix ranks, the allowed discrepency between these two can still be large compared with the total range of $[0,1]$.

\subsection{Connecting NBS and CKA through Uhlmann's theorem}
\label{subsec:cka-nbs-bounds-2}

The equality of the Procrustes and Bures distances can be further understood by noticing that for any particular fixed PSD $\mbK_X$, there are infinitely many matrices $\mbX$ for which $\mbK_X =  \mbX \mbX^{\top}$. This set of matrices are related by orthogonal transformations-- for two $M \times N_x$ matrices $\mbX$ and $\mbX'$ that satisfy $\mbX \mbX^{\top} = \mbK_X = \mbX' \mbX'^{\top}$, we have $\mbX \mbU = \mbX' $ with $\mbU^{\top} \mbU = \mbU \mbU^{\top} = \mbI$.
This set includes rectangular matrices of dimension $M \times N_x$, where $N_x \ge \text{rank}(\mbK_x)$.  
The unique PSD square root $\sqrt{\mbK_X}$ represents a particular square member of this set.

We will now assume $N_x = N_y = N$, so that we can compute the Hilbert-Schmidt inner product between $\mbX$ and $\mbY$ to measure their overlap. 
Intuitively, a meaningful measure of similarity between $\mbK_X$ and $\mbK_Y$ could be the maximum inner product over all $\mbX$ and $\mbY$ that are consistent with $\mbK_X$ and $\mbK_Y$.
Since these possible neural representations are all related by a orthogonal transformation, we can fix $\mbX$ and $\mbY$ arbitrarily and optimize their overlap over the set of orthogonal transformations $\mathcal{U}(N)$:
\begin{equation}\label{eq:Uhlmanns}
\begin{split}
    & \max_{\mbX, \mbY} \{ | \Tr[\mbX^{\top} \mbY] | : \mbX \mbX^{\top} = \mbK_X, \mbY \mbY^{\top} = \mbK_Y   \} \\ 
    & = \max_U \{ | \Tr [\mbX^{\top} \mbY \mbU ] | : \mbU \in \mathcal{U}(N) \} \\
    & = \| \mbX^{\top} \mbY \|_{*} = \mathcal{F}(\mbK_X, \mbK_Y) .
\end{split}
\end{equation}

In the last equality we have used the result of \cref{lemma:fidelity-nuclear-norm}.  
This result, known in the quantum information community as Uhlmann's theorem \cite{watrous2018}, shows that the fidelity between covariance matrices can be understood as the solution to maximizing the overlap between neural representation matrices that are consistent with those covariance matrices.
This result does not depend on the dimension $N$, provided it is at least the maximum rank of $\mbK_X$ and $\mbK_Y$ (adding extra dimensions beyond $\max(  \text{rank} (\mbK_X),\text{rank} (\mbK_Y) )$ does not affect the solution to this problem).
\Cref{eq:Uhlmanns} implies that NBS is simply the same maximum overlap between `consistent' neural representation matrices, normalized to lie in the interval $[0,1]$.  
Without loss of generality, we can write the maximization problem in \cref{eq:Uhlmanns} in terms of the unique $M \times M$ PSD square roots of the covariance matrices:

\begin{equation}\label{eq:fid-opt}
\mathcal{F}(\mbK_X, \mbK_Y) = \max_U \{ | \Tr [(\mbX \mbX^{\top})^{\frac{1}{2}} (\mbY \mbY^{\top})^{\frac{1}{2}} \mbU ] | : \mbU \in \mathcal{U}(M) \} 
\end{equation}

Choosing $\mbU = \mbI$ leads to the inequality:
\begin{equation}\label{eq:CKAinequal1}
    NBS(\mbK_X,\mbK_Y) \ge \frac{  \Tr [\mbK_X^{1/2} \mbK_Y^{1/2} ]  }{\sqrt{\Tr \mbK_X \Tr \mbK_Y}} = CKA(\mbK_X^{\frac{1}{2}},\mbK_Y^{\frac{1}{2}}).
\end{equation}
where we have dropped the absolute value signs because the Hilbert-Schmidt inner product between two PSD matrices is non-negative.  
The inequality is saturated when $\sqrt{\mbK_X}$ commutes with $\sqrt{\mbK_Y}$.  
This inequality shows that the CKA between two particular neural representation matrices, namely the PSD square roots $\sqrt{\mbK_X}$ and $\sqrt{\mbK_Y}$, appears as a suboptimal solution to the maximization in \cref{eq:fid-opt}.  
NBS($\mbK_X,\mbK_Y$), on the other hand, represents this same Hilber-Schmidt inner product maximized over all neural representation matrices consistent with $\mbK_X$ and $\mbK_Y$.

Lastly, we can derive an upper bound on NBS in terms of $CKA(\sqrt{\mbK_X},\sqrt{\mbK_X})$ using the Fuchs-van de Graaf inequalities from quantum information theory (see \cref{subsec:fuchs-graaf}).  
We are lead to another set of inequalities which bound the deviation of 
$CKA(\sqrt{\mbK_X},\sqrt{\mbK_X})$
from $NBS(\mbK_X,\mbK_Y)$:
\begin{equation}\label{eq:sqCKAbound}
     1 - NBS(\mbK_X,\mbK_X) \leq 1 - CKA(\mbK_X^{1/2},\mbK_Y^{1/2} ) \leq \sqrt{1 - NBS(\mbK_X,\mbK_X)^2}
\end{equation}
These upper and lower bounds are represented in \cref{fig:NBS-CKA} (b) as the orange and blue dashed lines.

%% file: 060-discussion.tex
\section{Discussion}
\label{sec:discussion}

Differences in neural representations are quantified by a wide variety of methods in the current literature~\cite{Klabunde2023}.
In many cases, the relationships between these various quantities is unclear both conceptually and quantitatively.
While prior works have made attempts to mathematically relate various (dis)similarity measures, our work greatly expands the scope of these comparisons.
In particular, we show that two independently proposed approaches---shape distances~\cite{kendall2009shape,Williams2021} and the normalized Bures similarity (NBS;~\cite{Muzellec2018,Tang2020})---are essentially identical (see \cref{theorem:duality}).
A notable feature of this equivalence is that the two methods are motivated from very different perspectives.
The Procrustes and Riemannian shape distances can be viewed as the residual distance that is left after neural dimensions are aligned by an optimal rotation.
In contrast, NBS directly compares the structure of two kernel matrices, $\mbK_X$ and $\mbK_Y$, without any alignment.
Superficially, NBS looks similar to CKA (see eq.~\ref{eq:CKA-NBS-matrix-norms}), but we have seen that these similarity scores utilize fundamentally different geometries (eqs.~\ref{eq:cka-euclidean-dist-connection} and~\ref{eq:nbs-bures-dist-connection}) and can produce discrepant quantitative outcomes as we demonstrated both analytically and numerically (see \cref{fig:NBS-CKA}).

NBS and Bures distance are rooted in a rich literature in quantum information theory~\cite{nielsen00,quantum-fidelity-measures-review,watrous2018}.  
Indeed, the bound on CKA derived in \cref{subsec:cka-nbs-bounds-2} follows a classic result in this area known as Uhlmann's theorem.
Similarly, the Bures geometry on PSD manifolds has been extensively studied in the context of optimal transport, yielding both theoretical insights and practical algorithms~\cite{chewi2020gradient,kroshnin2021statistical}.
Our work only scratches the surface of these connections, and future studies should seek to import additional findings from these well-developed nearby fields.

Our main result shows a duality between shape and Bures distances, and an important open question is whether similar dualities can be found for other (dis)similarity measures.
If shape and Bures distances represent a truly unique link between the two major perspectives on the problem (summarized in \cref{subsec:mappings,subsec:similarity-matrices}), this provides a concrete motivation for their adoption.
In short, these methods can enjoy the conceptual and practical advantages of each perspective, depending on the circumstance.

%% file: 99-SI.tex
\section{Proofs}


We use the following notation: if $\mbA$ is a positive semidefinite matrix, then it admits a decomposition $\mbA = \mbU \mbLambda \mbU^\top$ and a unique positive semidefinite square root $\mbA^{1/2} = \mbU \mbLambda^{1/2} \mbU^\top$.

\subsection{Proof of Lemma~\ref{lemma:procrustes-covariance}}
\label{subsec:proof-of-lemma-1}

\begin{proof}[Proof of \cref{lemma:procrustes-covariance}] We begin with the Procrustes distance:
\begin{align*}
\bigg [ \mathcal{P}(\mbX, \mbY) \bigg ]^2 &= \min_{\mbQ^\top \mbQ = \mbI} \Vert \mbC \mbX - \mbC \mbY \mbQ \Vert_F^2 \\
&= \min_{\mbQ^\top \mbQ = \mbI} \left ( \Tr[\mbX^\top \mbC^\top \mbC \mbX] + \Tr[\mbY^\top \mbC^\top \mbC \mbY] - 2 \Tr[\mbX^\top \mbC^\top \mbC \mbY \mbQ ] \right ) \\
&= \Tr[\mbX^\top \mbC \mbX] + \Tr[\mbY^\top \mbC \mbY] - 2 \max_{\mbQ^\top \mbQ = \mbI} \Tr \left [ \mbX^\top \mbC \mbY \mbQ \right ] \\
&= \Tr \left [ \mbSigma_{XX} \right ] + \Tr \left [ \mbSigma_{YY} \right ] - 2 \max_{\mbQ^\top \mbQ = \mbI} \Tr \left [ \mbSigma_{XY} \mbQ \right ] \\
&= \Tr \left [ \mbSigma_{XX} \right ] + \Tr \left [ \mbSigma_{YY} \right ] - 2 \Vert \mbSigma_{XY} \Vert_* 
\end{align*}

where the step in the last line is the well-known result of \textcite{Schonemann1966}. Similarly for the Riemannian shape distance:
\begin{align*}
\theta^*(\mbX, \mbY) &= \min_{\mbQ^\top \mbQ = \mbI} \arccos \left ( \frac{\Tr [\mbX^\top \mbC^\top \mbC \mbY \mbQ ]}{\sqrt{\Tr [ \mbX^\top \mbC^\top \mbC \mbX  ] \Tr [ \mbY^\top \mbC^\top \mbC \mbY  ] } } \right ) \\
&= \min_{\mbQ^\top \mbQ = \mbI} \arccos \left ( \frac{\Tr [\mbX^\top \mbC \mbY \mbQ ]}{\sqrt{\Tr [ \mbX^\top \mbC \mbX  ] \Tr [ \mbY^\top \mbC \mbY  ] } } \right ) \\
&=  \arccos \left ( \frac{ \max_{\mbQ^\top \mbQ = \mbI} \Tr [\mbSigma_{XY} \mbQ ]}{\sqrt{\Tr [ \mbSigma_{XX}   ] \Tr [ \mbSigma_{YY}   ] } } \right ) \\
&= \arccos \left ( \frac{ \Vert \mbSigma_{XY} \Vert_* }{\sqrt{\Tr [ \mbSigma_{XX}   ] \Tr [ \mbSigma_{YY}   ] } } \right )
\end{align*}

as claimed by the lemma.
\end{proof}

\subsection{Proof of Lemma \ref{lemma:fidelity-nuclear-norm}}
\label{subsec:proof-fidelity-nuclear-norm}
\begin{proof}
First, the nonzero singular values of $\mbX^\top \mbY$ are equal to the square root of the nonzero eigenvalues of $\mbA = \mbX^\top \mbY \mbY^\top \mbX$. Thus,
\begin{equation}
\label{eq:lemma1-eqa}
\Vert \mbX^\top \mbY \Vert_* = \Tr[(\mbX^\top \mbY \mbY^\top \mbX)^{1/2}] = \Tr[\mbA^{1/2}]
\end{equation}
Next, we argue that every nonzero eigenvalue of $\mbA \in \reals^{N \times N}$ is also an eigenvalue of ${\mbB = \mbX \mbX^\top \mbY \mbY^\top \in \reals^{M \times M}}$.
To see this, suppose $\lambda \neq 0$ is an eigenvalue of $\mbA$ with eigenvector $\mbv \in \reals^N$.
Then, $\mbw = \mbX \mbv \in \reals^M$ is an eigenvector of $\mbB$ with the same eigenvalue, since:
\begin{align}
\label{eq:lemma1-eqb}
\mbX^\top \mbY \mbY^\top \mbX \mbv &= \lambda \mbv  \\
\mbX \mbX^\top \mbY \mbY^\top \mbX \mbv &= \lambda \mbX \mbv &&\text{(multiply both sides on the left by $\mbX$.)}\\
\mbX \mbX^\top \mbY \mbY^\top \mbw &= \lambda \mbw &&\text{(let $\mbw = \mbX \mbv$.)}
\end{align}
Notice that \cref{eq:lemma1-eqb} together with $\lambda \neq 0$ implies that $\mbw = \mbX \mbv \neq \mb{0}$.
Further, $\mbB$ does not contain any additional nonzero eigenvalues or additional repeated eigenvalues, since:
\begin{align}
\text{rank}(\mbB) &= \text{rank}(\mbX \mbX^\top \mbY \mbY^\top)
\\
&\leq \text{rank}(\mbX^\top \mbY) && \text{(matrix product rank inequality)}
\\
&= \text{rank}(\mbX^\top \mbY \mbY^\top \mbX) && \text{(rank of a matrix and its Gram matrix are equal)}
\\
&= \text{rank}(\mbA).    
\end{align}
Thus, we've shown that the non-zero eigenvalues of $\mbA$ and $\mbB$ are equal.

Next, we define $\mbC = (\mbX \mbX^\top)^{1/2} \mbY \mbY^\top (\mbX \mbX^\top)^{1/2} \in \reals^{M \times M}$, and argue that it has the same eigenvalue spectrum as $\mbB$.
To see this, suppose that $\lambda \neq 0$ is an eigenvalue of $\mbC$ with eigenvector $\mbz \in \reals^M$.
Then, $\mbw = (\mbX \mbX^\top)^{1/2} \mbz$ is an eigenvector of $\mbB$ with the same eigenvalue:
\begin{align}
(\mbX \mbX^\top)^{1/2} \mbY \mbY^\top (\mbX \mbX^\top)^{1/2} \mbz &= \lambda \mbz  \\
\mbX \mbX^\top \mbY \mbY^\top  (\mbX \mbX^\top)^{1/2} \mbz &= \lambda  (\mbX \mbX^\top)^{1/2} \mbz &&\text{(multiply on the left by $(\mbX \mbX^\top)^{1/2}$.)}\\
\mbX \mbX^\top \mbY \mbY^\top \mbw &= \lambda \mbw &&\text{(let $\mbw = (\mbX \mbX^\top)^{1/2} \mbz$.)}
\end{align}

Combining this with our argument above, we conclude that the non-zero eigenvalues of $\mbA$, $\mbB$ and $\mbC$ are equal.
Furthermore, from the definition $\mbC$ and the definition of the fidelity in \cref{eq:fid-definition}, we have ${\Tr[\mbC^{1/2}] = \cF(\mbX \mbX^\top, \mbY \mbY^\top)}$.
We can therefore conclude the proof since, recalling \cref{eq:lemma1-eqa}, we have:
\begin{equation}
\Vert \mbX^\top \mbY \Vert_* = \Tr[\mbA^{1/2}] = \Tr[\mbB^{1/2}] = \Tr[\mbC^{1/2}] = \cF(\mbX \mbX^\top, \mbY \mbY^\top)
\end{equation}
as claimed by the lemma.
\end{proof}

\subsection{Applying the Fuchs-van de Graaf inequalities to NBS and CKA}\label{subsec:fuchs-graaf}

 One of the Fuchs-van de Graaf inequalities tells us how the fidelity bounds the nuclear norm of the difference between positive semidefinite matrices $\rho$ and $\sigma$ with trace equal to 1 (known as the trace distance) \cite{watrous2018}:
\begin{equation}
    \| \rho - \sigma   \|_{*} \leq 2 \sqrt{1 - \mathcal{F}(\rho,\sigma)^2}
\end{equation}

Rewriting $\rho = \mbK_X / \Tr[\mbK_X]$ and $\sigma = \mbK_Y / \Tr[\mbK_Y]$ allows us to recognize $\mathcal{F}(\rho,\sigma)$ as $NBS(\mbK_X,\mbK_Y)$.    
Using the norm inequality for positive semi definite operators $\mbA$ and $\mbB$, $\| \mbA - \mbB \|_{*} \ge \| \sqrt{\mbA} - \sqrt{\mbB} \|_F^2 $ and expanding, we have:

\begin{equation}
     1 - NBS(\mbK_X,\mbK_X) \leq 1 - CKA(\mbK_X^{1/2},\mbK_Y^{1/2} ) \leq \sqrt{1 - NBS(\mbK_X,\mbK_X)^2}
\end{equation}

where the first inequality on the left hand side is from \cref{eq:CKAinequal1}.  
The right hand side inequality can be rewritten 

\begin{equation}
    NBS(\mbK_X,\mbK_X) \leq \sqrt{1 - (1-CKA(\mbK_X^{1/2},\mbK_Y^{1/2} )^2 }.
\end{equation}

which defines the orange dashed curve in \cref{fig:NBS-CKA} (b).  